# Language Models and Retrieval Augmented Generation for Automated Structured Data Extraction from Diagnostic Reports: Assessment of Approaches and Parameters

Mohamed Sobhi Jabal [1]; Pranav Warman [2]; Jikai Zhang [3,4]; Kartikeye Gupta [5]; Ayush Jain [5]; Maciej Mazurowski [1,2,3]; Walter Wiggins [1]; Kirti Magudia [1]; Evan Calabrese [1,5]

**Author Affiliations:**
1. Department of Radiology, Duke University, Durham, NC, USA
2. Duke University School of Medicine, Durham, NC, USA
3. Department of Electrical and Computer Engineering, Duke University, Durham, NC, USA
4. Duke Center for Artificial Intelligence in Radiology, Duke University, Durham, NC, USA
5. Duke University, Durham, NC, USA
6. Department of Biomedical Engineering, Duke University, Durham, NC, USA

**Corresponding Author:**
Mohamed Sobhi Jabal
Department of Radiology
Duke University, Durham, NC
mohamedsobhi.jabal@duke.edu

**Abstract**

**Purpose**: To develop and evaluate an automated system for extracting structured clinical information from unstructured radiology and pathology reports using open-weights large language models (LMs) and retrieval augmented generation (RAG), and to assess the effects of model configuration variables on extraction performance.

**Methods and Materials**: The study utilized two datasets: 7,294 radiology reports annotated for Brain Tumor Reporting and Data System (BT-RADS) scores and 2,154 pathology reports annotated for isocitrate dehydrogenase (IDH) mutation status. An automated pipeline was developed to benchmark the performance of various LMs and RAG configurations. The impact of model size, quantization, prompting strategies, output formatting, and inference parameters was systematically evaluated.

**Results**: The best performing models achieved over 98% accuracy in extracting BT-RADS scores from radiology reports and over 90% for IDH mutation status extraction from pathology reports. The top model being medical fine-tuned llama3. Larger, newer, and domain fine-tuned models consistently outperformed older and smaller models. Model quantization had minimal impact on performance. Few-shot prompting significantly improved accuracy. RAG improved performance for complex pathology reports but not for shorter radiology reports.

**Conclusions**: Open LMs demonstrate significant potential for automated extraction of structured clinical data from unstructured clinical reports with local privacy-preserving application. Careful model selection, prompt engineering, and semi-automated optimization using annotated data are critical for optimal performance. These approaches could be reliable enough for practical use in research workflows, highlighting the potential for human-machine collaboration in healthcare data extraction.

**Introduction**

Healthcare reports including diagnostic radiology and pathology reports contain vital information necessary for patient care and various clinical and research purposes. Most remain unstructured and challenging to harness. Artificial intelligence (AI) and classical natural language processing (NLP) demonstrated some feasibility for automated data extraction from reports, but considerable challenges included computational cost, variable performance, and ungeneralizable task-specific models. More recently, transformer-based large language models (LMs) have emerged as the preeminent deep learning architecture for a variety of language processing tasks with wide range of potential healthcare applications (1–3). However, using leading commercial LMs requires transfer of data to private companies, which raises considerable privacy concerns for healthcare data and typically requires legal agreements and paid contracts (4–10).

Current development in open-weights LMs offers high performance while mitigating data privacy concerns. Recent studies proposed and demonstrated the potential of LMs and retrieval augmented generation (RAG) for extracting data from clinical documents (11–13). However, one major challenge is the large number of configuration variables to consider including the choice of model, model size, quantization level, embedding models, retrieval strategies, prompting methods, sampling parameters, and output formatting. As each of these factors may affect model performance and computational requirements, comprehensive evaluation of their effects on clinical data extraction tasks is valuable for effective use of these approaches.

This study presents an automated system designed to extract structured clinical information from unstructured healthcare reports using state-of-the-art open-weights LMs and RAG. We examined the LM pipeline configuration influence on data extraction effectiveness using radiology and pathology report datasets from patients with malignant brain tumors. By presenting a structured investigation across various LM architectures and retrieval strategies, this work aims to elucidate recommended strategies for implementing automated clinical data extraction systems from healthcare reports.

## Materials and Methods

### Dataset and Preprocessing

The study utilized two healthcare report datasets from the same single-institution, retrospectively identified cohort of 1,490 patients undergoing treatment for malignant brain gliomas (adult-type diffuse gliomas, WHO grades 2-4 based on pathologic diagnosis). The radiology dataset consisted of 7,294 radiology reports for brain MRIs with and without contrast performed for brain tumor follow-up, which typically include Brain Tumor Reporting and Data System (BT-RADS) scores at our institution (14,15). Reports were manually annotated for the reported BT-RADS score [0, 1, 1a, 1b, 2, 2a, 2b, 3, 3a, 3b, 3c, 4, or unprovided]. The pathology dataset consisted of 2,154 surgical pathology reports for brain tumor resections including molecular analyses. Reports were manually annotated for the reported isocitrate dehydrogenase (IDH) mutation status [positive - mutant, negative - wildtype, or unprovided]. Non-canonical IDH and IDH-2 mutations were considered "mutant". Dataset labeling was accomplished using custom, task-specific, rules-based NLP algorithms and subsequently manually reviewed and corrected by a neuroradiologist ([REDACTED for peer review]). The dataset was preprocessed using regex and standard NLP techniques to clean and normalize the text, filtering out newline characters unfollowed by a period and ensuring spaces surrounding newlines after periods, maintaining readability and sentence separation.

### Pipeline, Optimization, and Evaluation

We developed an automated pipeline for systematic performance assessment of various LM configuration variables. The benchmarking pipeline was applied on subsets of 500 randomly selected reports from the radiology and pathology datasets. Models were categorized into two groups based on size and computational demand representing high- and low-resource settings: smaller models having ≤ 8B parameters and ≤ 5-bit quantization. The best performing models and configurations were selected for subsequent evaluation on the full dataset. Performance was assessed using accuracy, micro and macro F1, precision, and recall. Independent samples t-test, Welch's t-test assuming unequal variances, and paired t-test were conducted as applicable. Correlations were examined using Spearman's rank coefficient. All pipeline testing was performed on a workstation equipped with an NVIDIA A6000 48 GB GPU. A generalized adaptation of the report processing pipeline for assessing model configurations is provided at [GitHub/WebApp].

**Language Models**

The study employed state-of-the-art open-weights LMs to assess their performance in extracting clinical information from the radiology reports (16). The LMs used varied in size, quantization level, training data, release date, medical fine-tuning, and they included Llama3, openbiollm-Llama3, Llama2, and MedIlama2, Meditron, Mistral, Biomistral, Mixtral, Phi3. Given its recent release, Llama3.1 was applied post-hoc to the entire datasets.

**Model Parameters and Quantization**

To evaluate the trade-off between model parameter size and performance, we tested a range of small and large parameter-size models. We tested a range of available quantization degrees from 3-bit to 16-bit according to the available levels for each model up to 48 GB VRAM. Quantization reduces memory and computational needs of language models by using fewer bits to represent parameters. This allows LMs deployment on GPUs with smaller VRAM, at some cost of precision loss.

**Prompting Strategies**

We evaluated the performance of simple and complex prompting strategies, with and without few-shot prompting. Simple prompting involved a single sentence describing the task, while complex prompting provided detailed instructions on valid responses and output format. Few-shot prompting included examples of correct data extraction and negative examples to handle non-reported data. Models were instructed to format responses as a JSON object schema.

**Sampling Parameters**

The pipeline systematically evaluated various inference-time sampling parameters' influence on extraction accuracy, specifically: temperature, top-k, and top-p sampling. Temperature, which controls the randomness and indeterministic nature of the model's outputs, was varied across a range with approximate logarithmic steps [0.0, 0.01, 0.1, 0.5, 0.8]. Top-k sampling, which limits the number of tokens considered and samples from the k most likely next tokens, was investigated at the values: [2, 5, 10, 40]. Top-p sampling controls the cumulative probability cutoff distribution of considered tokens and was investigated at [0.1, 0.5, 0.9].

**Output Formatting**

To examine effects of enforced standardization of model outputs on performance, we implemented a JSON output formatting from Ollama. This structured output format was compared against prompt instruction to evaluate its extraction accuracy impact.

**Retrieval and Embedding Approaches**

The pipeline was assessed with and without RAG. The following RAG steps were applied: recursive character text splitting to divide the report into chunks based on sentence separators. Chunk size was set at 70 with 20-character overlap. Embedding model "gte-large" was implemented. Processing was orchestrated using Langchain framework. Text embeddings were stored and indexed in a local vector database (FAISS, Meta AI). Dense retrieval using cosine similarity of vector embeddings was applied to process extracted RAG context chunks, additionally two other optional retrieval approaches were included: hybrid ensemble retrieval combining BM25 with dense retrieval and sequential retrieval using the two methods consecutively. BM25 is a widely used keyword-based search and ranking algorithm that has been essential for many modern search engines (17). A cross-encoder reranker (bge-reranker-v2-m3 from Beijing Academy of Artificial Intelligence) was utilized to sort the retrieved chunks and prioritize the most relevant ones. Keywords relevant to the targeted data point were "follow-up score" and "IDH IDH1 IDH2 IDH1/IDH2 detected positive negative" for radiology and pathology reports, respectively. The retrieved chunk with the highest relevance score was selected, with a relevance score threshold of 0.2, below which RAG was disused.

**Postprocessing**

Custom postprocessing was applied to the raw language model outputs to enhance consistency and validity. The target datapoint was first extracted, addressing formatting variations such as delimiters. Artifacts like newline characters, extra whitespace, and inconsistent string quotation marks were removed. Any extraneous text surrounding the desired JSON-formatted output was discarded, and the cleaned response was parsed as a JSON object, retaining only the first valid entry if multiple were present. The extracted information was then validated against the predefined set of valid values for both datasets, with unexpected formats and hallucination edge cases, including null or malformed responses, defaulting to invalid. This postprocessing ensured the final data was clean, consistent, and ready for downstream evaluation.

**Results**

The performance of 13 language models in 407 different configurations was evaluated on automatically extracting BT-RADS follow-up scores from radiology reports and IDH mutation status from pathology reports. An overview of the pipeline design is provided as Figure 1. Mean radiology report word count was 265 (± 66) with a median of 255 (IQR: 80) words, and the mean pathology report word count was 2504 (± 2563) with a median of 1218 (IQR: 3721) words. The reference standard distribution for BT-RADS follow-up scores in the radiology dataset were: 1: 5 (0.07%), 1a: 204 (2.80%), 1b: 124 (1.70%), 2: 856 (11.74%), 2a: 112 (1.54%), 2b: 10 (0.14%), 3: 88 (1.21%), 3a: 47 (0.64%), 3b: 292 (4.00%), 3c: 386 (5.29%), 4: 373 (5.11%), NR: 4,797 (65.77%). As for IDH mutation reference standard pathology reports, the class distribution was: Positive: 154 (7.15%), Negative: 1559 (72.38%), Not Provided: 441 (20.47%). Sampled examples of the analyzed radiology and pathology reports are provided in Figure 2.

**Benchmarking Model Performance**

Overall, benchmarking performance varied significantly across different models, configurations, and approaches. On the radiology sampled dataset (n=500), accuracy ranged from 62.6% to 98.4% with a mean of 88.4% (±8.9%) and median of 88.8% (IQR: 9.8%). As for the pathology reports, accuracy ranged from 5.2% to 93.6% with a mean of 57.7% (±27.1%) and median of 50.9% (IQR: 45.7%).

Ordered by accuracy, the top performing smaller models were (Acc, M-F1): openbiollm-llama-3 8B Q_4 (99.2%, 98.8), mistral 7B Q_4 (99.0%, 97.9), phi3 4B Q_4 (99.0%, 90.2), llama3 8B

Q_4 (98.4%, 82.4), and biomistral 7B Q_4 (98.0%, 88.0). While the top benchmarked larger models were llama3 70B Q_3 (99.2%, 98.8), mixtral 56B Q_4 (99.2%, 98.8), mistral 7B Q_6 (99.0%, 90.5), openbiollm-llama-3 8B Q_6 (95.2%, 85.5), and phi3 14B Q_5 (93.6%, 82.7).

**Model Parameters and Quantization**

Larger models demonstrated higher mean accuracy (86% ± 22%) compared to smaller ones (75% ± 32%) (t = 3.80, p < 0.001), with medium effect (Cohen's d = 0.40). In terms of model parameter size, there was significant positive correlation with accuracy (r = 0.25, p < 0.001) and macro F1 (r = 0.18, p < 0.001). Quantization level correlations with performance were statistically non-significant for accuracy (r = 0.06, p = 0.280) and macro F1 (r = 0.08, p = 0.134). Illustration of the relationship and correlations between model performance, parameter size, quantization, and recency are demonstrated in Figure 3.

**Prompting Strategies**

In the radiology dataset, complex prompting consistently improved accuracy across models (32.2% to 98.6% vs. 18.6% to 92.6%). The mean accuracy increase was 12.01% ± 11.56% ($t$=3.11, $p$=0.01). Similarly, few-shot prompting improved accuracy in most models. Mean accuracy with few-shot prompting ranged from 21 to 96% vs. 0 to 98.6% without it ($t$=3.00, $p$=0.02), the mean increase being 32.42% ± 32.39%. When including negative examples in the few-shots method of reports lacking the datapoint of interest, accuracy improved for most models, with accuracy ranging from 34.6-98.0% vs. 16.81-92.61% without negative examples ($t$=2.01, $p$=0.08), and an average accuracy increase of 10.65% ± 15.89%. Effect of different prompting methods on performance is illustrated in Figure 3.

**Sampling Parameters**

Regarding temperature, correlations across all metrics were small, ranging from 0.009 to 0.026 ($p$=0.78-0.91), without statistically significant impact on performance. Similarly, top-p and top-k parameters also had minimal correlation of (0.01-0.04, p=0.76-0.91) and (0.01-0.03, $p$=0.75-0.91), respectively. Model performance across varying temperature, top-k, top-p is shown in Figure 6.

**Output Formatting**

In the radiology dataset, enforcing JSON output generally improved accuracy (range with JSON: 12.0%-99.15% vs. without: 20.53%-92.57%, mean change being +2.92% ± 6.78%), without statistical significance ($t$=1.29, $p$=0.23). Contrastively, the pathology dataset showed minimal, inconsistent effects from JSON output (range with JSON: 37.3%-69.3% vs. without: 37.3%-68.3%; average change: -0.44% ± 1.54%) ($t$=-0.64, $p$=0.56). Effect of forcing JSON output on model accuracy is demonstrated in Figure 5.

**Retrieval Augmented Generation**

In the radiology sample set, the influence of RAG on accuracy varied widely. The mean accuracy with RAG enabled ranged from 24.0% to 90.8%, compared to 19.73% to 94.35% without RAG. Mean accuracy change was -7.75% ± 31.10% ($t$=-0.91, $p$=0.39). Contrastively in the pathology sample set, RAG consistently improved accuracy across all models. The mean accuracy with RAG ranged from 69.4% to 92.8% vs. 5.2% to 44.8% without RAG, a statistically significant mean increase of 48.08% ± 11.18% ($t$=7.91, $p$=0.001). RAG impact on language model performance is shown in Figure 5.

**Global Model Performance**

The performance of the best model configurations evaluated on the entire radiology (n=7,294) and pathology (n=2,154) datasets were as follows (Accuracy, Macro F1): On the full radiology dataset, the top 5 results ordered by accuracy were: openbiollm-llama-3 70B Q_4 (98.68%, 90.96%), llama3 70B Q_3 (98.66%, 90.95%), llama3.1 70B Q_4 (98.64%, 90.93%), llama3.1 8B Q_4 (96.74%, 77.80%), phi3 4B Q_4 (94.98%, 87.27%). And for the complete pathology dataset: openbiollm-llama-3 70B Q_4 (90.02%, 85.81%), llama3 70B Q_3 (88.58%, 84.96%), llama3.1 70B Q_4 (85.93%, 75.36%), llama3.1 8B Q_4 (80.22%, 71.97%), phi3 4B Q_4 (76.79%, 65.92%). The overall results are presented in Table 1 and Table 2, respectively.

**Discussion**

In this study, we developed an automated system for structured clinical data extraction from unstructured radiology and pathology reports using open-weights LMs and RAG approaches.

The top performing models were: openbiollm-llama3:70B, llama3:70B, and llama3.1:70B, achieving over 98% accuracy for radiology BT-RADS score extraction and reaching 90% for pathology IDH mutation status extraction across the entire datasets. These findings demonstrate the technology's significant potential to accurately identify and fetch key clinical information from complex diagnostic reports with minimal human intervention and suggest the models could be reliable enough for practical use in research workflows and automated curation of structured databases from unstructured reports.

Results from our configuration experiments provide potentially helpful insights for designing similar systems for structured data extraction from healthcare reports: First, larger and more recent models consistently and measurably outperformed their counterparts across different settings, conferring an advantage on extraction tasks. A notable exception was Phi:3.8B, a small recent LM performed on par with large models, suggesting model architecture and training can partially compensate for smaller parameter size, which is appealing for local on-device efficient deployment in resource-constrained environments with limited computation capacity. Medically fine-tuned llama3 was the best performing model overall (none was available for llama3.1 at the time of the study). Of note, despite its non-leading performance, llama3.1 features significantly longer context window and improved multilingual capabilities compared to its predecessor. Second, we observed minor performance impact of model quantization across the 3-16-bit range, suggesting VRAM-constrained use-cases should ideally prioritize larger models with low-bit quantization. Third, prompting strategy had significant impact on performance, specifically the use of complex detailed prompts and inclusion of few-shot prompting resulting in notable accuracy increase, underscoring the importance of prompt engineering. Fourth and finally, we found that several inference-time and sampling parameters (temperature, top p, top k) had little or no effect on performance including model.

This study builds upon recent work demonstrating the potential of LMs for healthcare data extraction. Prior studies have shown LMs can extract RADS features with 86-99% accuracy (18) and deduce various categorization from radiology reports with 75% accuracy (19), which focused primarily on small datasets (30-160 reports) and private commercial LMs regardless of output formatting. Huang et al. assessed LM-based structured data extraction from clinical notes using ChatGPT-3.5 achieving an accuracy of 89% in extracting classifications from 1000 lung cancer pathology reports (20). Glicksberg et al. noted positive observations using GPT-4 for predicting emergency admissions from initial clinical variables with RAG resulting in increased AUC (0.79 to 0.87) (21). Most prior studies focused on commercial LMs, a notable exception was a recent study by Guellec et al., examining open LM (Vicuna) performance in extracting data from 2,398 radiology reports and achieving accuracies > 95%, with unstructured outputs (22).

The study adds to this body of work by exploring recent open LM models and configurations to examine and help identify recommendable parameters across compute capabilities. It further explored the potential impact of RAG and found negative performance effect on relatively short and simple radiology reports and substantial positive performance impact on longer (10-fold) complex pathology reports. The main advantage of RAG lies in increasing signal-to-noise ratio by retrieving only the most relevant sentences, which was especially helpful in reports where the target datapoint (e.g., IDH mutation status) is mentioned multiple times in different contexts throughout the document.

Another key insight was the value of using a semi-automated approach with sampled annotated data for benchmarking and pipeline optimization before full dataset application. This is particularly helpful given the large number of tunable parameters associated with local custom LM systems, given LMs' sensitivity to minor prompt variations, output formatting, and jailbreaking, significantly altering performance (23).

This highlights the crucial interplay between domain expertise and automated AI-systems in healthcare and demonstrates the advantage of partially automated systems in improving accuracy, monitoring, and reliability. At the current phase, we believe human-machine collaboration is the most valid approach for data extraction and research is the most relevant use case. Error rates are expected to drop in the future, which may ultimately enable human independent implementation and clinical use cases. Recent integration of structured output feature from leading closed LM providers (24) indicates possible future adoption from open frameworks potentially increasing the reliability of valid JSON output generation.

*Limitations*

Our findings should be considered within the scope and limitations of the study; The radiology and pathology datasets, while relatively large, had fairly homogeneous reporting structure given the single center source. In addition, the chosen target variables had discrete categorical responses, and performance could decrease for more ambiguous datapoints. Finally, while we explored a large configuration space, it was impractical to be exhaustive.

## Conclusion

The study highlights the significant potential of open LMs for local automated extraction of structured clinical data from radiology and pathology reports. We found that larger, newer, and domain fine-tuned models had higher performance, that model quantization had minimal effect, that prompt engineering and methods is critical for high performance, and that several inference and sampling parameters had little to no effect. RAG significantly improved performance in complex and lengthy pathology reports. Depending on report complexity, top-performing model configurations achieved 90% to 98% accuracy in structured extracting of clinical datapoints underscoring the approach reliability for practical use in research workflows. Semi-automated systems with human oversight and iterative optimization using partially annotated data proved important when developing LM systems.

# Tables

**Table 1**. Language model performance on the entire radiology report dataset.

| Language Model | Parameters | Quantization | Accuracy | Macro Precision | Micro Precision | Macro Recall | Micro Recall | Macro F1 | Micro F1 |
|---|---|---|---|---|---|---|---|---|---|
| Openbiollm-llama3 | 70 | 4 | 98.68% | 90.48 | 98.68 | 91.45 | 98.68 | 90.96 | 98.68 |
| Llama3 | 70 | 3 | 98.66% | 90.48 | 98.66 | 91.44 | 98.66 | 90.95 | 98.66 |
| Llama3.1 | 70 | 4 | 98.64% | 90.52 | 98.64 | 91.35 | 98.64 | 90.93 | 98.64 |
| Llama3.1 | 8 | 4 | 96.74% | 75.92 | 96.74 | 84.73 | 96.74 | 77.8 | 96.74 |
| Llama3 | 8 | 4 | 96.22% | 75.55 | 96.22 | 84.77 | 96.22 | 77.86 | 96.22 |
| Phi3 | 4 | 4 | 94.98% | 84.78 | 94.98 | 90.86 | 94.98 | 87.27 | 94.98 |
| Biomistral | 7 | 4 | 86.5% | 66.51 | 86.5 | 84.06 | 86.5 | 70.32 | 86.5 |
| Mixtral | 56 | 8 | 45.19% | 90.27 | 45.19 | 67.01 | 45.19 | 74.43 | 45.19 |

**Table 2**. Language model performance on the entire pathology report dataset.

| Language Model | Parameters | Quantization | Accuracy | Macro Precision | Micro Precision | Macro Recall | Micro Recall | Macro F1 | Micro F1 |
|---|---|---|---|---|---|---|---|---|---|
| Openbiollm-llama-3 | 70 | 4 | 90.02% | 82.47 | 90.02 | 90.38 | 90.02 | 85.81 | 90.02 |
| Llama3 | 70 | 3 | 88.58% | 82.08 | 88.58 | 89.44 | 88.58 | 84.96 | 88.58 |
| Llama3.1 | 70 | 4 | 85.93% | 77.19 | 85.93 | 77.25 | 85.93 | 75.36 | 85.93 |
| Mixtral | 56 | 8 | 84.35% | 63.13 | 84.35 | 63.34 | 84.35 | 62.69 | 84.35 |
| Llama3.1 | 8 | 4 | 80.22% | 72.58 | 80.22 | 76.33 | 80.22 | 71.97 | 80.22 |
| Biomistral | 7 | 4 | 77.48% | 49.14 | 77.48 | 51.71 | 77.48 | 45.93 | 77.48 |
| Phi3 | 4 | 4 | 76.79% | 62.76 | 76.79 | 77.1 | 76.79 | 65.92 | 76.79 |
| Llama3 | 8 | 4 | 42.25% | 66.17 | 42.25 | 68.25 | 42.25 | 52.74 | 42.25 |

**Figures**

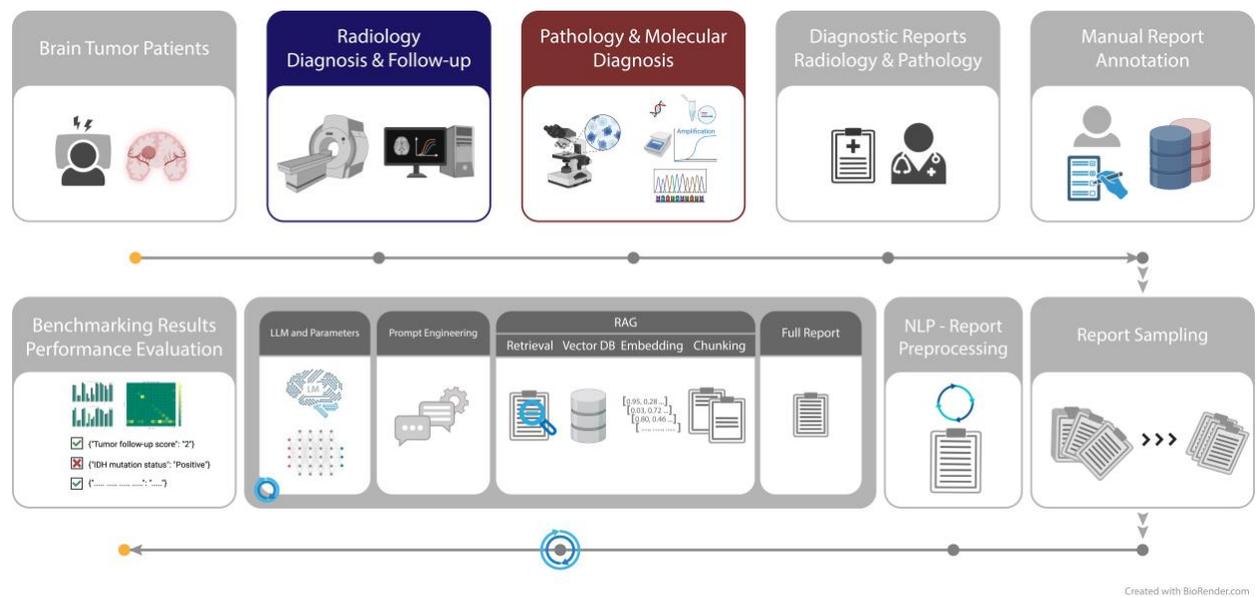

**Figure 1.** Overview of the study workflow demonstrating the different steps from diagnosis to automated structured data extraction of relevant follow-up and clinical information from diagnostic radiology and pathology reports.

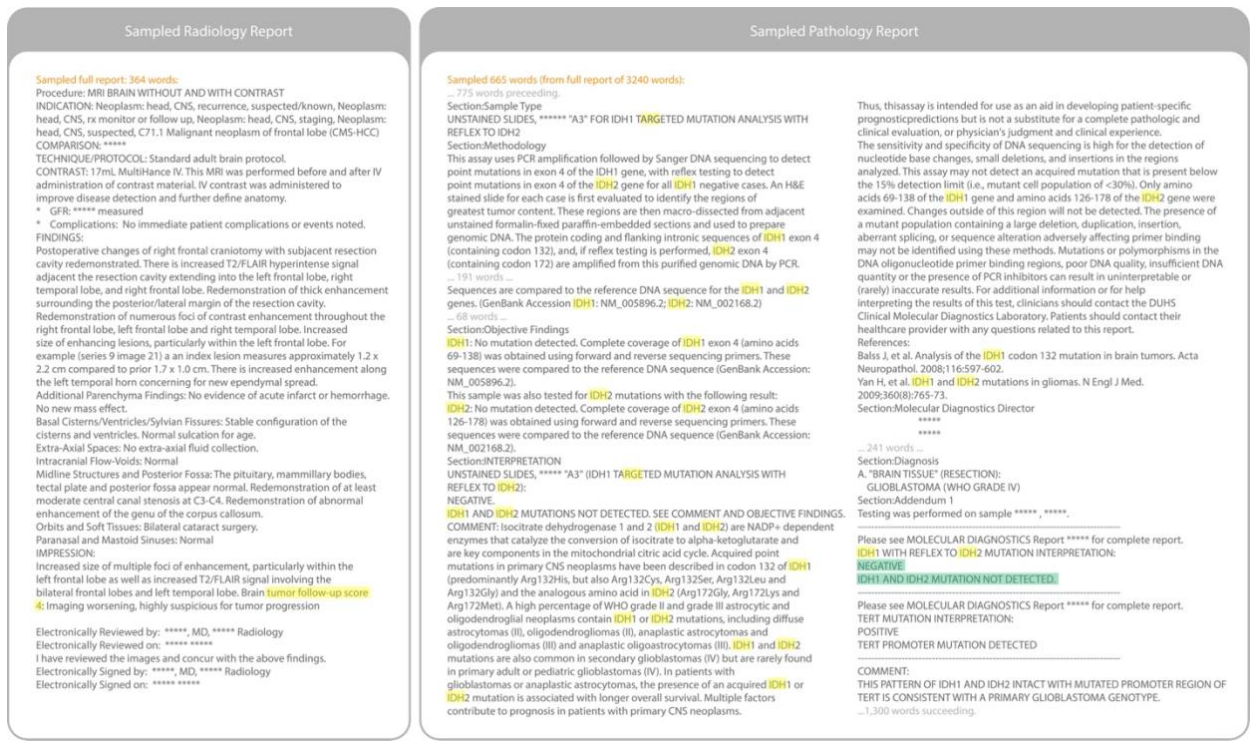

**Figure 2**. Sampled examples of processed radiology and pathology reports with highlighted datapoints of interest in yellow and retrieved context using RAG in green.

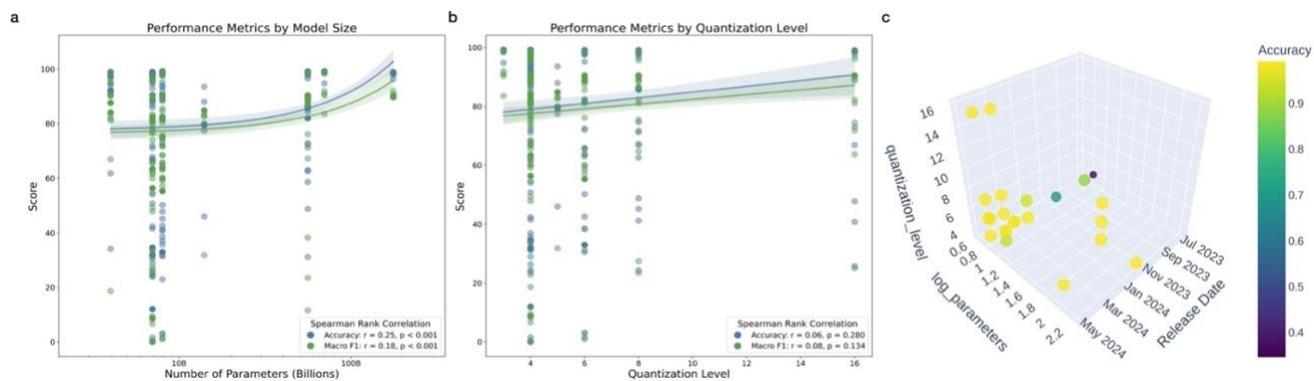

**Figure 3.** Correlation plots showing the relationships between model performance metrics (accuracy and macro F1 score) versus logarithm of model parameter size and model quantization level (a, b). Size and release date distribution of deployed language models with correlated performance for structured clinical data extraction from radiology reports (c).

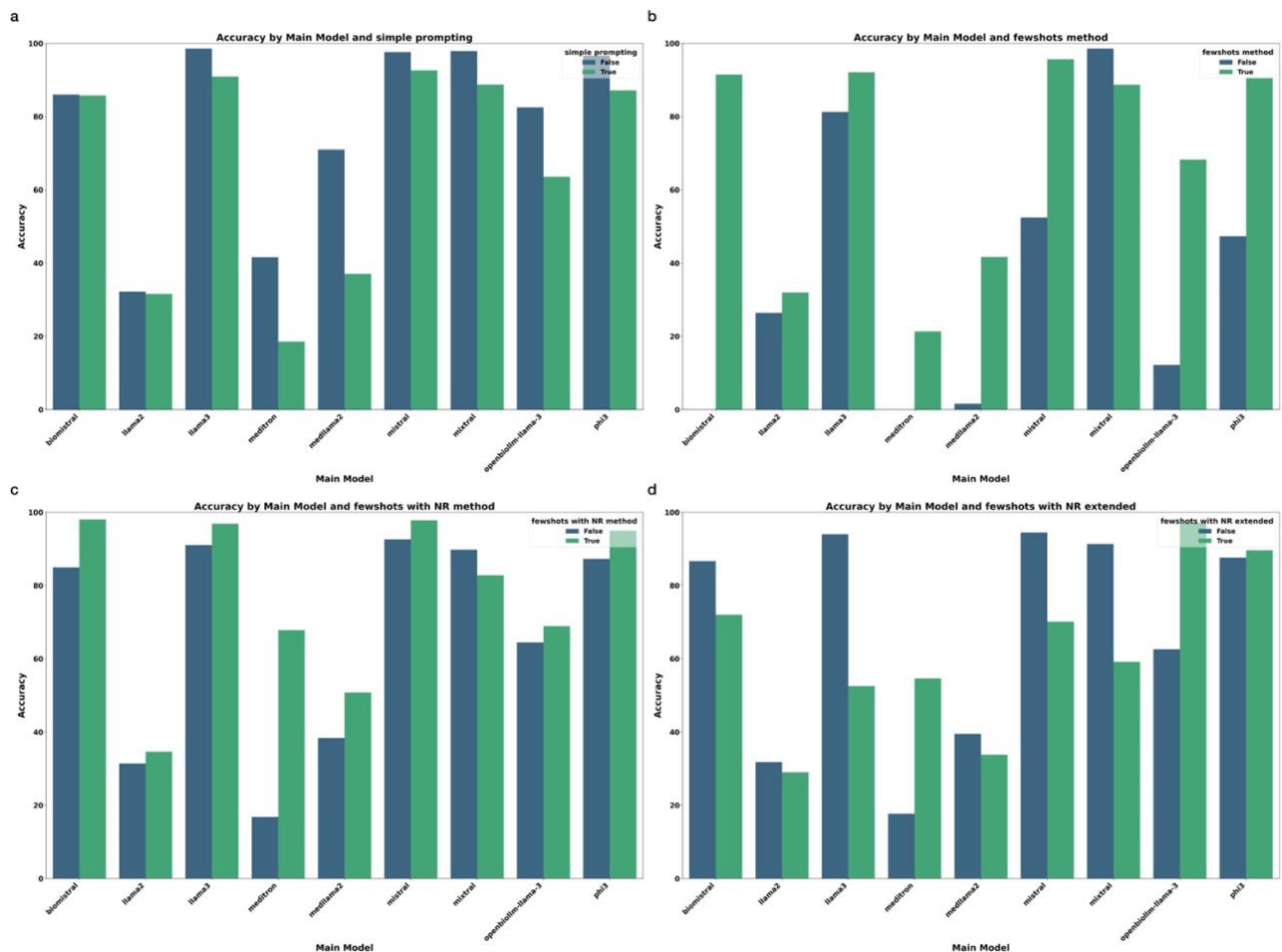

**Figure 4.** Language model performance comparing simple and complex detailed prompting approaches (a) and fewshots methods with positive (b) and negative non-reported (NR) examples (c, d) for structured clinical data extraction from radiology reports.

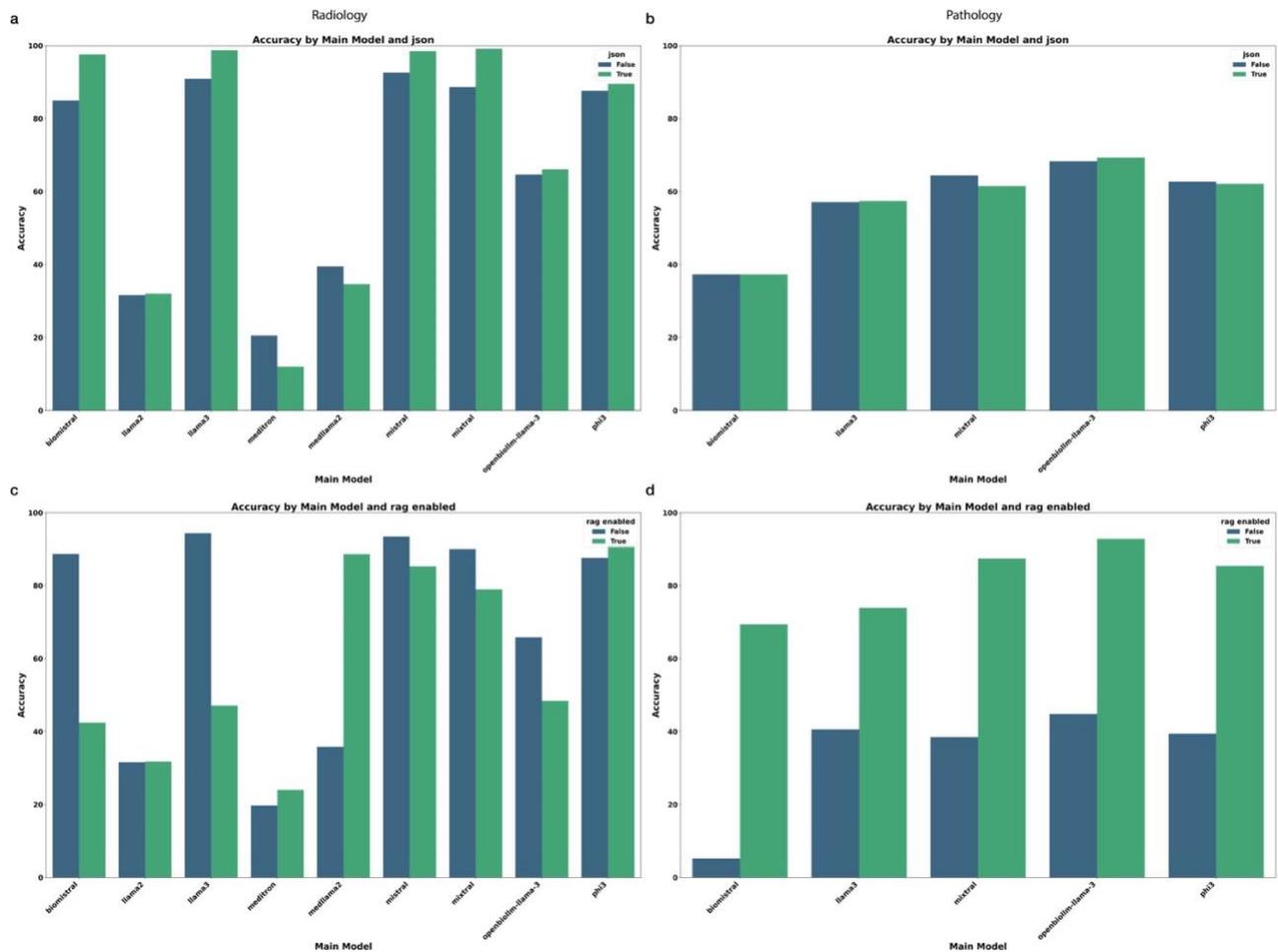

**Figure 5.** Language model performance evaluating the effect of forcing JSON outputs and adopting RAG for structured clinical data extraction from radiology (a, c) and pathology (b, d) reports.

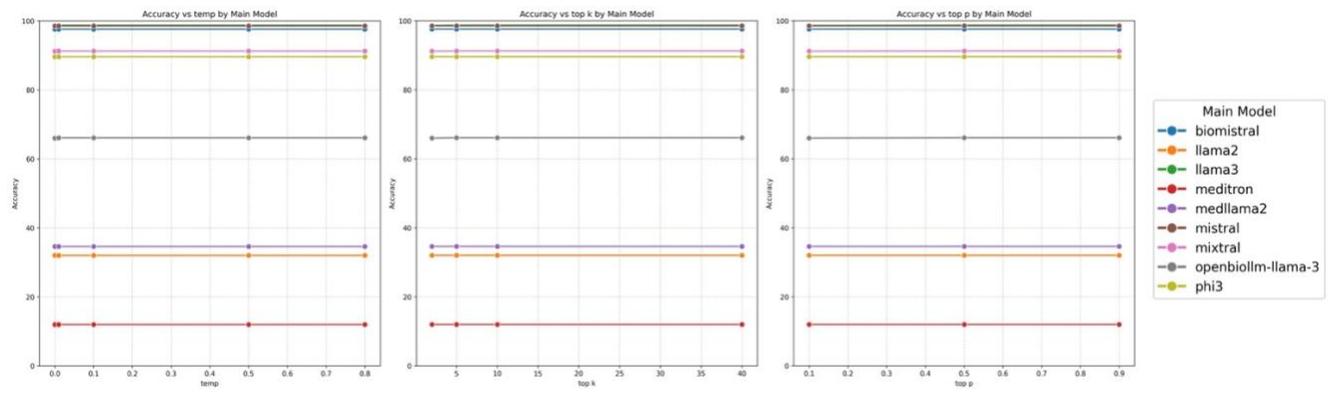

**Figure 6.** Language model performance across varying inference-time parameters, temperature, top p, and top k for structured clinical data extraction from radiology reports.